\documentclass[10pt]{article} 
\usepackage[preprint]{rlc}

\usepackage{amssymb}            
\usepackage{mathtools}          
\usepackage{mathrsfs}           
\mathtoolsset{showonlyrefs}     
\usepackage{graphicx}           
\usepackage{subcaption}         
\usepackage[space]{grffile}     
\usepackage{url}                

\title{A Mixture-of-Experts Approach to Few-Shot Task Transfer in Open-Ended Text Worlds}

\author{Christopher Z. Cui  \\
    ccui46@gatech.edu \\
    Georgia Institute of Technology
    \And
    Xiangyu Peng  \\
    xpeng62@gatech.edu \\
    Georgia Institute of Technology
    \And
    Mark O. Riedl  \\
    riedl@cc.gatech.edu \\
    Georgia Institute of Technology}

\begin{document}

\maketitle

\begin{abstract}

Open-ended worlds are those in which there are no pre-specified goals or environmental reward signal.
As a consequence, an agent must know how to perform a multitude of tasks.
However, when a new task is presented to an agent, we expect it to be able to reuse some of what it knows from previous tasks to rapidly learn that new task.
We introduce a novel technique whereby policies for different {\em a priori} known tasks are combined into a Mixture-of-Experts model with an attention mechanism across a mix of frozen and unfrozen experts.
The model learns when to attend to frozen task-specific experts when appropriate and learns new experts to handle novel situations.
We work in an open-ended text-based environment in which the agent is tasked with behaving like different types of character roles and must rapidly learn behaviors associated with new character role types. 
We show that our agent both obtains more rewards in the zero-shot setting, and discovers these rewards with greater sample efficiency in the few-shot learning settings.


\end{abstract}

\section{Introduction}
\label{sec:intro}

Open-ended worlds are those in which there are no pre-specified goals or environmental reward signal.
In such environments, we might want to have an agent that is capable of carrying out a variety of different behaviors and achieving a variety of goals based on the what is needed.
In this work we look at the challenge of task-transfer in open-ended environments.
In an open-ended environment, an agent must know how to perform a multitude of tasks.
In this setting, we assume an agent has learned one or more policies, each for a specific task, but now must acquire a new policy for a new task.
Policy models that perform distinct tasks can be gathered into a Mixture of Experts (MoE) where an attention mechanism learns which expert to listen to when performing actions.
We show that the MoE can learn new tasks in a few-shot fashion when there are aspects of the new task that result behaviors from the experts.

We work in an open-ended text environment based loosely on Dungeons \& Dragons.
Text environments are those in which an agent receives natural-language descriptions of its immediate locale in the environment, and performs actions by describing its actions with text. 
Text-based worlds have become a benchmark challenge for reinforcement learning agents~\citep{hausknecht2020interactive, narasimhan2015language, ammanabrolu2019playing, ammanabrolu2019graph, ammanabrolu2020avoid, adhikari2020learning,ALFWorld20,murugesan2020enhancing, Wang2022Science, peng2023story, pan2023rewards, prasad2023adapt, abdulhai2023lmrl, carta2023grounding, wang2023behavior, ryu-etal-2023-minimal, Prateek2023}.
They are exemplified by the classical game, {\em Zork}, in which an agent must solve puzzles to get to the end of the game. 
Text-based games are challenging for the following reasons~\citep{hausknecht2020interactive}:
First, they are partially-observable environments; what can be observed is usually limited to a ``room’’.
Second, they have very large action spaces.
For example, the game, {\em Zork} can accept commands up to four words in length and has a vocabulary of 700, meaning that there are $700^4$ possible actions available in each state.
Not all word sequences create meaningful change to the environment.
Third, commonsense and trope knowledge is usually required to complete games.
Action sequences that make sense in the real world usually also make sense in text games (e.g., ``open mailbox’’ vs ``eat mailbox’’) and have similar effects, though science fiction and fantasy tropes may also be present (e.g., silver bullets kill vampires).
Fourth, unlike many computer games, text-games often involve solving puzzles with long-horizon causal and temporal relations.

Most commercial text games---such as those in the {\em Jericho} benchmark suite~\citep{hausknecht2020interactive}---have an objective to solve, and thus a reward signal can be constructed around completion of the game and/or 
progress through the game. 
On the other hands, Table-top role-playing games such as Dungeons \& Dragons are open-ended worlds in the sense that one can go nearly anywhere and do nearly anything.
Often there is a quest or mission, but that quest or mission is not pre-ordained and often unknown in advance.
A character may have a particular role---fighter, thief, etc.---which is accompanied by particular behavioral expectations.

These behavioral expectations can be used to frame acting in a particlar role as a task. 
That is, absent a mission or quest, an agent has a task of role-playing in character.
If a mission or quest is presented, then the task is to accomplish the goal by incorporating behavior expectations into their actions (a hunter may be more interested in making progress by engaging in combat whereas a thief may be more interested in progress via stealth and trickery). 

From this, we can define the challenge of task transfer in role-playing games as a single agent that must transfer knowledge from one or more known character roles to a new character role type. 
%
Transfer is feasible because, although a role such as a thief will perform many actions that  a hunter would never perform, the thief will also need to handle common situations such as navigating through the starting town, though perhaps visiting separate locations or interacting with the townsfolk in a different manner.
As is often the case in task transfer in text games, some of an original policy is applicable to the new task without adaptation, but not all of it.
By assembing expert, role-specific policy models into a Mixture of Experts, one can learn a new role faster because one of the experts may be able to make an informed guess about what the agent should do in the new context.

We introduce a novel technique whereby $N$ frozen experts in the MoE model propose actions in the next task, and attention is applied across experts. 
An $N+1^\mathrm{th}$, untrained ``expert'' is added to the ensemble but remains hot so that it can learn from interactions with the environment.
This ``hot'' expert learns how to interact with the environment when all the other frozen experts are insufficient, and the MoE as a whole learns the new task more rapidly than training a new expert from scratch or fine-tuning existing experts.
This result holds even when the new task is not a strict blend of existing experts.

\section{Related Work}
\label{sec:related}

\subsection{Text Adventure Game Playing Agents}

Text based games have shown great potential for use as Reinforcement Learning benchmark environments~\citep{hausknecht2020interactive, narasimhan2015language}. 
\citet{ammanabrolu2019playing} 
proposed augmenting reinforcement learning with knowledge graphs as external memory about world state.
\citet{ammanabrolu2019graph} proposed KG-A2C, which integrates knowledge graphs into the actor-critic~\citep{bahdanau2016actor} RL framework.
The Q*BERT agent~\citep{ammanabrolu2020avoid} further extended KG-A2C to incorporate the BERT~\citep{devlin-etal-2019-bert} language model into the model architecture.
We build on top of the KG-A2C family of models since they have shown state-of-the-art performance.
Other techniques for playing textgames include GATA~\citep{adhikari2020learning}, which builds a knowledge-graph based  representation of the world on top of a transformer-based agent, training through a combination of RL and self-supervised learning.



Whereas text adventure games have pre-defined progression toward a goal state, table-top role playing games involve open-ended game play.
We refer to text-based environments that support open-ended game play as {\em text-based role playing} to signify the interaction with the environment through reading and writing text instead of verbal interactions with other players and game masters.

The LIGHT environment~\citep{urbanek2019learning} is a crowdsourced text-based role playing game with a 
rich environment with interactable NPCs, objects and locations, each with a short paragraph description, demonstrating the value of grounding in training agents that can not only act but also converse successfully. 
\citet{ammanabrolu2021motivate} propose agents that can switch seamlessly between generating natural language and action declarations. 
These agents can learn to play different characters when given a motivation that includes character type and goal as part of the world state.

{\em Story Shaping}~\citep{peng2023story} is a technique for training RL agents to play text role-playing games wherein a story is converted into a rich reward signal. 
The technique can be used to train different characters, but can only train a single agent to emulate a single character.
Our character-based reward strategy is related, but our rewards are manually crafted instead of inferred from stories.

\subsection{Ensembling in RL}

To our knowledge, transfer learning with a mixture of pre-trained experts for RL has not been explored but there exists a large body of literature on transfer learning and ensemble methods in RL. Prior works explore ensemble methods of using multiple, frozen experts (teachers) to train a new agent as a student where the goal is to minimize the difference between the teachers' and student's policies ~\citep{hinton2015distilling, rusu2016policy, Yin2017, schmitt2018kickstarting, schulman2018equivalence, Distral2017, parisotto2016actormimic}. Other approaches instead make direct use of the teachers' policies, either with some heuristic to evaluate the 'goodness' of the teacher's action with respect to the current state or via a learned discriminator ~\citep{Cheng2020, kurenkov2020ac, Fang2021, li2023iob}. Some works further explore methods for compensating for sub-optimal teachers ~\citep{kurenkov2020ac, Faulkner2021}. Attention modules have been utilized in RL before, but only with experts that are learned through segmenting features or inputs at training time in a multi-task setting ~\citep{sodhani2021multitask, Cheng2023}.
Our work resides in the intersection of these ideas. We leverage multiple, frozen sub-optimal experts in a MoE, using an attention module to combine the outputs of the experts. Rather than a student, an additional hot expert learns to bridge the gap when the sub-optimality of the pre-trained experts would leave the MoE otherwise unable to proceed. 

\subsection{Few-shot Adaption}
Large pre-trained Language models have emerged as extremely powerful tools for NLP tasks~\citep{devlin2019bert, raffel2020exploring, brown2020language}. However, a limitation of these powerful models is their size, some with parameters numbering in the billions \citep{brown2020language}. This makes them prohibitively expensive when it comes to further training or fine-tuning. 
Low-Rank Adaptation (LoRA) circumvents this by keeping the model frozen and introducing trainable rank decomposition matrices~\citep{hu2021lora}.
Our proposed technique also freezes the core model(s) and trains additional layers on top, though the specific mechanics needed for reinforcement learning are different.

\section{Background}
\label{sec:background}
\subsection{Text-Adventure Games}
\label{sec:TAG}
A text-adventure or text-based role playing game can be modeled as a partially-observable Markov decision process (POMDP) M = $\langle S, T, A, \omega, O, R, \gamma \rangle$ where
$S$ is the set of ground truth world states,
$A$ is the set of actions,
$T$ is the probability of transitioning from one state to another given an executed action,
$R$ is a reward function,
$O$ is the set of possible observations,
$\omega$ is the probability of observations given the ground truth world state, and
$\gamma$ is a parameter estimating the reward horizon~\citep{hausknecht2020interactive}.
In our setting, we will use a deterministic transition function $T$, which is common in text-based games. 
However, nothing in our proposed technique strictly requires it.
The objective of reinforcement learning is to learn a policy, $\pi:S \rightarrow A$ that maps states to actions, such that taking the action mapped to the current state and following the policy henceforth maximizes expected reward. 

\subsection{KG-A2C and Story Shaping}
\label{sec:kga2c}
We consider the standard reinforcement learning setting
where an agent interacts with a text game environment over a number of discrete time steps.
State-of-the-art approaches to RL in text environments use a knowledge graph as external, persistent memory of the world state~\citep{ammanabrolu2018playing, ammanabrolu2019graph, ammanabrolu2020avoid}.
As the agent explores the game world, a knowledge graph is constructed and used as state representation.
In this paper, we choose KG-A2C agent framework~\citep{ammanabrolu2019graph} as our reinforcement learning agent.

KG-A2C's space of observations, $o_t$,
includes 
(a)~text description of the room the agent is in, 
(b)~text descriptions of the character's inventory,
(c)~the agent's last command, and
(d)~feedback from the last command.
These state observations are processed using a GRU based encoder using the hidden state from the previous step, combining them into a single observation embedding.
Simultaneously, the state observation is used to update a knowledge graph $G_t$ of persistent memory of the world state.
This includes facts and relations about rooms, objects in rooms, inventory items, etc.
This knowledge graph is then embedded using a graph attention mechanism~\citep{veličković2018graph}.
The overall representation vector is
updated as $\mathbf{o}_{t}$.
The agent is trained via the Advantage Actor Critic (A2C) \citep{mnih2016asynchronous} method to maximize long term expected reward in the game in a manner otherwise unchanged from \citet{ammanabrolu2020avoid}.
More details about this agent can be found in Appendix \ref{KG-A2C}.


Agents output a language string into the game to describe the actions that they want to perform. 
To ensure tractability, this action space can be simplified down into templates.
Templates consist of interchangeable verbs phrases ($VP$), optionally followed by prepositional phrases ($VP$ $PP$), e.g. $([carry/take]$ \underline{\hspace*{.3cm}}$)$ and $([throw/discard/put]$ \underline{\hspace*{.3cm}} $[against/on/down]$ \underline{\hspace*{.3cm}}$)$, where the verbs and prepositions within $[.]$ are aliases.
Actions are constructed from templates by filling in the template's blanks using objects from the game's vocabulary.

In order to train different role-aligned policy models $\pi_i$, we follow the {\em Story Shaping} technique~\citep{peng2023story}.
Different stories are first converted into knowledge graphs $G_i$.
As the agent explores the game world, a knowledge graph---called the {\em World KG}---is constructed and used as state representation $G_t$.
The similarity between $G_i$ and $G_t$ is used to provide an intrinsic reward for the KG-A2C model, subsequently shaping the RL-Agent to conform to the specified roles.
More details about how we train KG-A2C with story shaping method can be found in Appendix \ref{Story Shaping}.

\subsection{LIGHT Environment}
\label{sec:LIGHT}
In our paper, we create a game in LIGHT environment~\citep{urbanek2019learning}, which provides
a text world environment with a database of 1775 Non-Player Characters (NPCs), 663 locations, and 3462 objects with rich text descriptions.
We've developed an open-world game where players can play various roles, each with its own set of activities. 
For example, the player/agent can play a ``\texttt{adventurer}'' in our game, and numerous activities have been specifically tailored for this role, including the presence of dragons specifically designed for `adventurers' to confront and defeat.

\section{Method}

\label{sec:method}

\label{sec:method}
\begin{figure}[tbh!]
    \raisebox{20pt}[\dimexpr\height+0\baselineskip\relax]{\includegraphics[width=\textwidth]{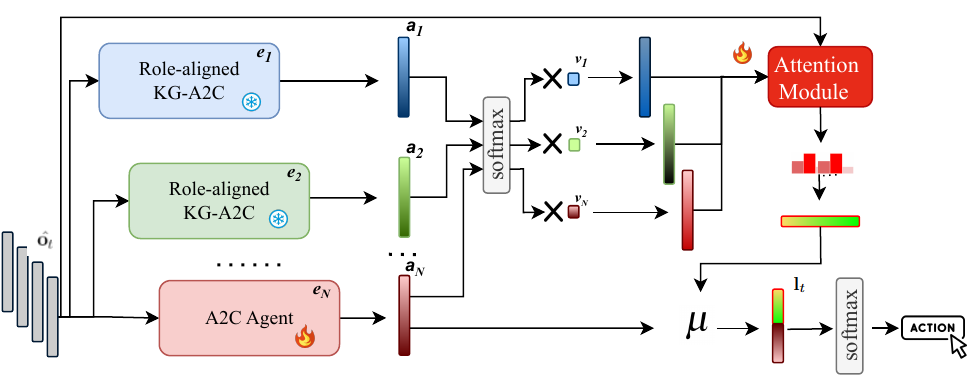}}%
    \centering
    \caption{Pipeline of our MoE Agent. At each time-step, all experts produces a logit distribution over actions, $\mathbf{a}_1, \mathbf{a}_2, ..., \mathbf{a}_N$. These are each passed through a softmax to get the resulting probabilities and multiplied by the $\mathbf{v}_1, \mathbf{v}_2, ..., \mathbf{v}_N$ produced by each expert's critic module. The scaled probabilities are then mixed by the attention module and averaged (operation represented by $\mu$) with the $\mathbf{a}_N$ produced by the trainable expert to obtain $\mathbf{l_t}$. These averaged logits are then passed through a softmax and sampled to produce an action.} 
    \label{fig:pipeline}
\end{figure}

Given a collection of pre-trained, role-aligned expert models $e_1, e_2, ...e_{N-1}$ from source characters $p_1, p_2,..., p_{N-1}$ pre-trained using story shaping methods~\citep{peng2023story},
our agent aims to use their knowledge on a new role task under few-shot settings.

\paragraph{Action Probabilities} 
We first initialize an empty expert $e_{T}$ and add it to the mixture-of-experts. 
This empty expert is a repository for any policy that is unique to the new task role and not sufficiently handled by another, existing expert.
We obtain a probability distribution over actions at time $t$ for each expert, $\mathbf{a}_
{t,1}, \mathbf{a}_{t,2}, ..., \mathbf{a}_{t,N}$,
where $ \mathbf{a}_{t,j} \in \mathbb{R}^{v}$, by feeding 
the overall representation vector $\mathbf{o}_{t}$ into frozen, pre-trained character-aligned agents, $e_1, e_2, ...e_{N-1}$ and trainable policy model $e_{N}$.
$v$ is the action space of the game environment.

The KG-A2C splits action generation into a multi-step process. A template distribution $\mathbf{\tau}_{t,j}$ is generated from $\mathbf{o}_{t}$ and is sampled to select template $T_\mathrm{act}$. Both of these are then used to generate a probability distribution over all game objects. A mask is applied to reduce this distribution to only admissible objects for the current state $\mathbf{O}_{t,j}$ and sampled to product an object, $T_\mathrm{obj}$. The resulting action $\mathbf{a}_{t,j}$ is a combination of $T_\mathrm{act}$ and $T_\mathrm{obj}$.

In our MoE agent, the template $T_\mathrm{act}$ sampled by the original model is used to generate the corresponding object distribution $\mathbf{O}_{t,j}$. This means for each pre-trained, character-aligned model $e_1, e_2, ...e_{N-1}$ that results in action templates $\mathbf{\tau}_{t, 1}, \mathbf{\tau}_{t, 2}, ..., \mathbf{\tau}_{t, \mathrm{N-1}}$, the templates sampled from these distributions are used to generate $\mathbf{O}_{t, 1}, \mathbf{O}_{t, 2}, ..., \mathbf{O}_{t, N-1}$. We represent both $\mathbf{\tau}_{t,j}$ and $\mathbf{O}_{t,j}$ with $\mathbf{a}_{t,j}$ going forward.

\paragraph{Value Seeding.} 
Each expert $e_1, e_2, ...,e_{N}$ possesses a critic $\mathbf{C}_1, \mathbf{C}_2, ..., \mathbf{C}_{N}$ that predicts an expected return $\mathbf{v}_{t, 1}, \mathbf{v}_{t, 2}, ..., \mathbf{v}_{t, N}$ given state S. These values represent how rewarding each expert believes their current state to be, or equivalently, the importance of the state to the respective, pre-trained model. These values can be used to infuse the MoE model with knowledge on which pre-trained expert to place more weight on at any given state due to an expectation of a higher return. We do this by multiplying each $\mathbf{a}_{t,j}$ with the corresponding $\mathbf{v}_{t,j}$ prior to being input to the attention module.
The critic modules are not explicitly shown in Figure~\ref{fig:pipeline}.

\paragraph{Expert Attention.}
An expert attention module $\mathcal{G}$ 
merges the output action distributions from each expert module.
It learns weights based on which expert to attend to and produces a final action probability distribution.
We use two separate attention modules, $\mathcal{G}_\mathrm{act}$ and $\mathcal{G}_\mathrm{obj}$, to account for the KG-A2C's multi-step action generation. However, these attention modules are identical outside of being modified to process the different sized inputs and will be collectively referred to as $\mathcal{G}$ going forward. The approach outlined in the following section can be generalized to any architecture with a single-step action generation. 

The input of $\mathcal{G}$ is the representation of observations $\hat{\mathbf{o}}_t$ in time step $t$ and action probabilities $[\mathbf{a}_
{t,1}; \mathbf{a}_{t,2}; ...; \mathbf{a}_{t,N}]$. We believe the distribution of action probabilities multiplied by the corresponding expected returns $\mathbf{v}_{1, j}, \mathbf{v}_{2, j}, ..., \mathbf{v}_{N, j}$ represent the knowledge possessed by the experts in our game environment; the expected returns represent how valuable each model perceives its current state to be. 
Multiplying the former by the latter and mixing the experts’ action probabilities is a good way to use this knowledge and save training time under few-shot setting. 

We first apply a feed-forward network to project $\hat{\mathbf{o}}_t$ non-linearly to a new representational space:
\begin{align}
    &\mathbf{h}_{t} = \mathrm{LayerNorm}(\mathbf{W}_{\mathrm{up},o}^\top \cdot\gamma(\mathbf{W}_{\mathrm{down},o}^\top\cdot\hat{\mathbf{o}}_t))
\end{align}
where $\gamma(\cdot)$ is a non-linear activation function.
We used SiLU~\citep{elfwing2018sigmoid} in our architecture.
$\mathbf{W}_{\mathrm{down},o} \in \mathbb{R}^{l \times l'}$ and $\mathbf{W}_{\mathrm{up},o} \in \mathbb{R}^{l' \times l}$ are projection parameters to be updated during
training.
Then we apply Layer Norm \citep{ba2016layer} to get $\mathbf{h}_{t} \in \mathbb{R}^{l}$ --- the final projected representation of observations $\mathbf{o}_t$.
Similarly, we project each action probability $\mathbf{a}_{t, j}\in \mathbb{R}^{v}$ of the role-aligned $c_j$ given observations $\mathbf{o}_t$ into the same space:
\begin{align}
    &\mathbf{h}_{l,j} = \mathrm{LayerNorm}(\mathbf{W}_{\mathrm{up},l}^\top \cdot\gamma(\mathbf{W}_{\mathrm{down},l}^\top\cdot\mathbf{a}_{t,j}))
\end{align}
where $\mathbf{W}_{\mathrm{down},l} \in \mathbb{R}^{v \times v'}$ and $\mathbf{W}_{\mathrm{up},l} \in \mathbb{R}^{v' \times v}$ are projection parameters to be updated during
training.

We compute the attentions by calculating the product between $\mathbf{h}_{{l,j}}$ and $\mathbf{h}_{o}$ and apply softmax over the experts, as follows:
\begin{align}
    \alpha_{t,j}=\frac{e^{(\mathbf{h}_{l,j} \cdot\mathbf{h}_{o})}}{\sum_{k=1}^{T} e^{(\mathbf{h}_{l,k}\cdot \mathbf{h}_{o})}}.
\end{align}
We then obtain the output logits $\mathbf{a}_{t} \in \mathbb{R}^{ v}$ by computing a linear combination of  $[\mathbf{a}_{t,1} * \mathbf{v}_{t,1}; ...; \mathbf{a}_{t,N} * \mathbf{v}_{t,N}] $ given the computed input-logit attention, 
\begin{align}
    \mathbf{l}_{t} &= \mathcal{G}(\hat{\mathbf{o}}, [\mathbf{a}_{t,1}; ...; \mathbf{a}_{t,N}]) = (\mathbf{a}_{t,N} + \sum_{j=1}^{N} \alpha_{t,j} \mathbf{a}_{t,j} )/2
    \label{eq:attention}
\end{align}
Then we follow the standard approach of Reinforcement learning agents, where the combined logits $\mathbf{l}_{t}$ are processed by second softmax transformation and then used to sample the action. This sampling is done in two stages where the output of $\mathcal{G}_\mathrm{act}$, $\mathbf{\tau}_{t,N}$, is sampled and used to generate $\mathbf{O}_{t,N}$.

\paragraph{Exploration.} 
Our agent tends towards premature convergence. Due to the Value Seeding we apply to the probabilities prior to the attention module, the probability mass of the resulting distribution post softmax tends to be most heavily weighted on the top action of the expert whose critic predicts the highest value. When this action results in a reward, this becomes a positive feedback loop where the attention module continuously places more weight on the expert associated with this reward until all other experts are ignored. This results in a pre-mature convergence on the portion of the reward function associated with the over-fit expert. This also leads to expert starvation, where an expert whose rewards do not appear until later in the environment are always assigned an extremely low weight by the attention module due to higher weights being assigned to experts whose rewards are found early.  

We implement two measures to solve this and drive exploration. 
We use (1)~Epsilon-Greedy sampling with decay instead of multinomial sampling, and (2)~implement a new loss term. 
Epsilon-Greedy sampling helps with exploration because the original KG-A2C uses an entropy loss to push the agent to explore and prevent the agent from prematurely convergence. 
While this is sufficient for training any single expert KG-A2C module, it is not sufficient to prevent the attention module from premature convergence. This in turn pushes the entire MoE agent towards 
prematurely converging on the scores associated with a specific pre-trained role. 
Epsilon-Greedy avoids this by sampling a new action in a manner that is influenced by neither the attention module nor the underlying A2C.

We add a new loss term to the KG-A2C loss.
This loss is calculated as the log sum of the action logits with the object portion of the loss normalized across the number of decoded objects. Let $\mathbf{l}_\mathrm{act}$ and $\mathbf{l}_\mathrm{obj}$ be the outputs of $\mathcal{G}_a$ and $\mathcal{G}_o$ respectively, where $\mathbf{\pi}_\mathrm{act}(\hat{\mathbf{o}}_t) = \{\tau_0, \tau_1, ... , \tau_K\}$ and $\mathbf{\pi}_\mathrm{obj}(\hat{\mathbf{o}}_t) = \{\mathbf{O}_0, \mathbf{O}_1, ... , \mathbf{O}_M\}$
represent all valid template and objects for a given state and $n$ is the number of objects used in $T_\mathrm{act}$:
$$\mathcal{L}(\mathbf{o}_t;\theta_t) = \frac{1}{K}\sum_{i=1}^K (1-\log(\tau_i | \mathbf{o}_t)) + \frac{1}{n}\sum_{p=1}^n\frac{1}{M}\sum_{j=1}^M (1-\log(\mathbf{O}_j | \mathbf{o}_t))$$
Functionally, this a cross entropy loss that flattens the action probability distribution by penalizing any particular action the closer its probability is to 1 and motivating a probability distribution that is more uniformly distributed between the most probable actions. 

\section{Experiments}
\label{sec:exp}
\paragraph{Experts.} We use as experts four KG-A2C models which have each been trained to emulate a role: 
$e_\mathrm{thief}$, $e_\mathrm{adventurer}$, 
$e_\mathrm{hunter}$, and $e_\mathrm{hoarder}$. The parameters of these experts are frozen. Another untrained KG-A2C model acts as the empty expert,  $e_\mathrm{hot}$, and attention modules $\mathcal{G}_\mathrm{act}$ and $\mathcal{G}_\mathrm{obj}$ are randomly initialized and trained during the few-shot finetuning on the game with a new role. 

\paragraph{Environment Details.} We execute the MoE agent and our baselines in the same open-ended environment that has multiple opportunities for actions that align with various personas. The environment (see Figure~\ref{app:game_map} in the Appendix) has a common starting room and an exit room that terminates the game when the agent enters it. 
Reward is earned when certain role-specific behaviors are performed in particular locations.
Some roles have more opportunities for role-aligned actions near the start room. Closer to the exit room there is a branch such that one role prefers one branch with the other roless preferring the other branch.

\paragraph{Baselines.} We test our MoE agent against two sets of baselines. All models used are from the KG-A2C~\citep{ammanabrolu2019graph} family of models and are trained with Story-Shaping~\citep{peng2023story}.
The first baseline is a KG-A2C agent trained from scratch on the new task.
We also use individual pre-trained expert policies, separated from the MoE and fine-tuned on the new role.
%
We train one \textit{Fine-tuned Expert} for each expert that goes into the full MoE agent.

\paragraph{Target Roles.} 
We evaluate our MoE agent across a number of different target roles. 
\textit{Blends} are target roles  composed of behaviors drawn from a subset of original four roles.
\textit{Partial blends} are target roles that are a mix of behaviors drawn from the original four roles but also completely new behaviors.
All target character are crafted to have roughly the same maximum scalar reward as the pre-trained experts were pre-trained on, and all target roles must reach the goal location. 
We show the results from a blended target role that has a roughly equal mix of behaviors from all previous roles (but is not a strict union of all rewards from all roles) and a partial blend target role that requires behaviors from only some pre-trained experts. Results from other target roles are available in the appendix. 

\subsection{Results}
\label{sec:results}

\begin{figure}[tbh!]
\includegraphics[width = \textwidth]{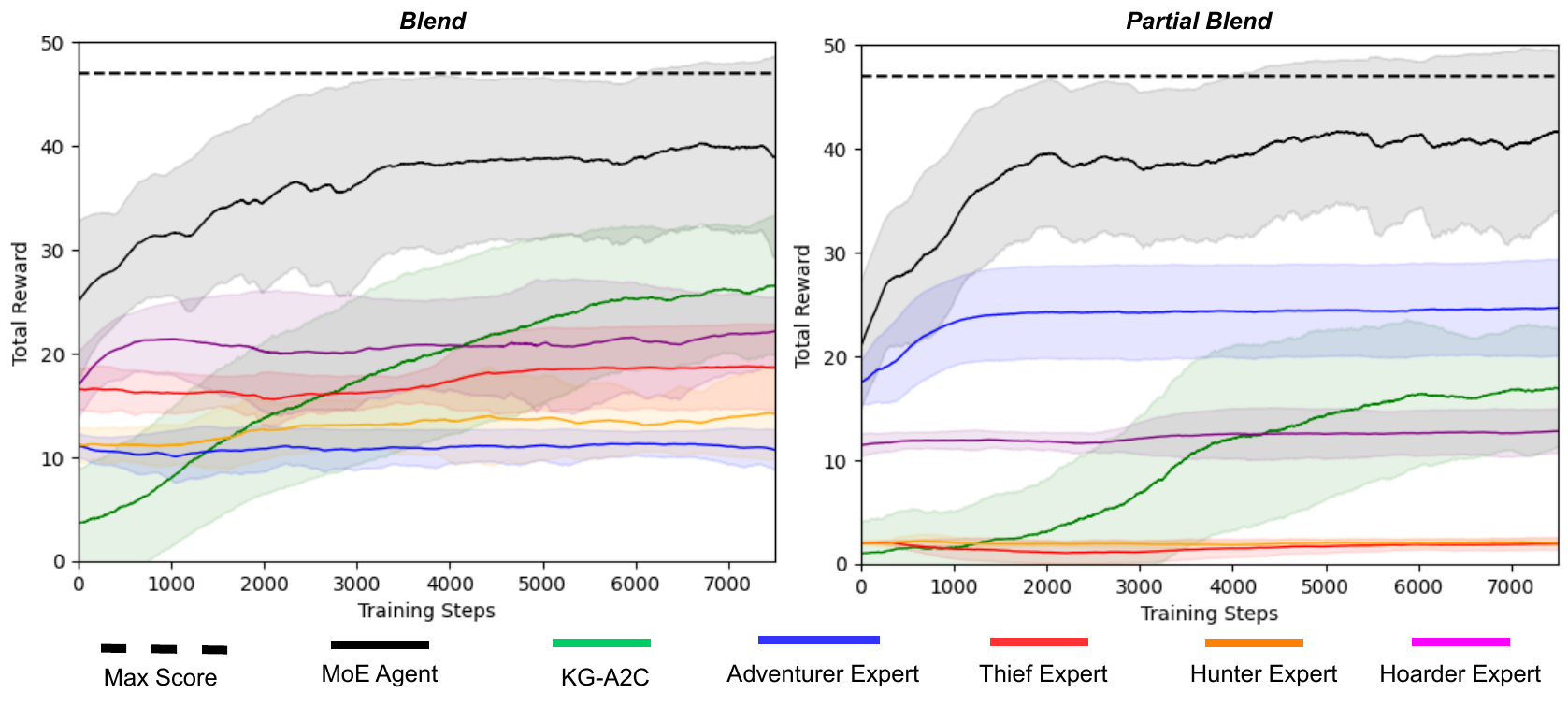}
    \caption{The full agent versus unfrozen experts allowed to continue training and a new KG-A2C. The left graph illustrates the test time performance on a persona composed only of pre-existing expert behaviors while the right illustrates the performance on a persona that is a mix of both pre-existing and new behaviors.}
    \label{fig:exp}
\end{figure}

The performance of the MoE agent can be measured through two metrics: sample efficiency and total score. Figure \ref{fig:exp} illustrates the training steps and testing time score of the full MoE agent against our baselines. We show the averaged results across 5 different seeds. Additional details and results are available in the Appendix.

\paragraph{Sample Efficiency.} While the MoE Agent produces a set of action probabilities for every expert, these distributions are generated from only one step in the environment. While this can result in an increase in training time compared to the baselines, the MoE agent has far greater sample efficiency and increases in score faster than the \textit{new KG-A2C} baseline. Conversely, the \textit{Fine-tuned Experts} all see little to no change in testing-time performance in the span of the training.

\paragraph{Total Score.}  
In both target roles, most \textit{Fine-tuned Experts} fail to find any additional rewards outside of those associated with their pre-trained roles and the small reward for reaching the goal room. This is especially evident on the partial blend target roles, where the pre-trained experts whose roles did not contribute to the target role achieved only the small reward for navigating to the goal location.
The \textit{New KG-A2C} discovers some rewards closer to the starting location over the course of the training period but fails to obtain rewards deeper in the environment. 
From testing time performance at training step 0 and after a short period of training, we see the MoE agent obtains a higher score in both a zero-shot setting due to the value seeding and a few-shot setting over time. The large standard deviation of the MoE agent is largely an artifact of the bias we place on exploration: across 20 random seeds over all target roles, only one seed failed to produce a model that failed to achieve the max score when evaluated over 10 random seeds.



\section{Expert Composition Study}
\label{sec: able}


In this section, we look at whether the composition of experts matter, with regard to having more experts and less, and the extent to which the relevance of experts helps or hurts.
The first test examines the robustness of the MoE agent for $e_1, e_2, ...e_{\mathrm{N-1}}$ as $N$ increases. 
The second examines the performance of the MoE agent in the absence of any experts relevant to the target role. 

\paragraph{Distractor Experts.} To examine the MoE agent's robustness to a large number of irrelevant experts, we double the number of pre-trained, frozen experts by adding 4 randomly initialized KG-A2C models. These new models are all kept frozen as with all other pre-trained experts. We follow the same training and evaluation procedure and show the results for the same roles used in Section \ref{sec:results}. For visual clarity, we only plot the performance of the regular MoE agent against the same agent with four additional, randomly initialized experts.

\begin{figure}[]
\includegraphics[width = \textwidth]{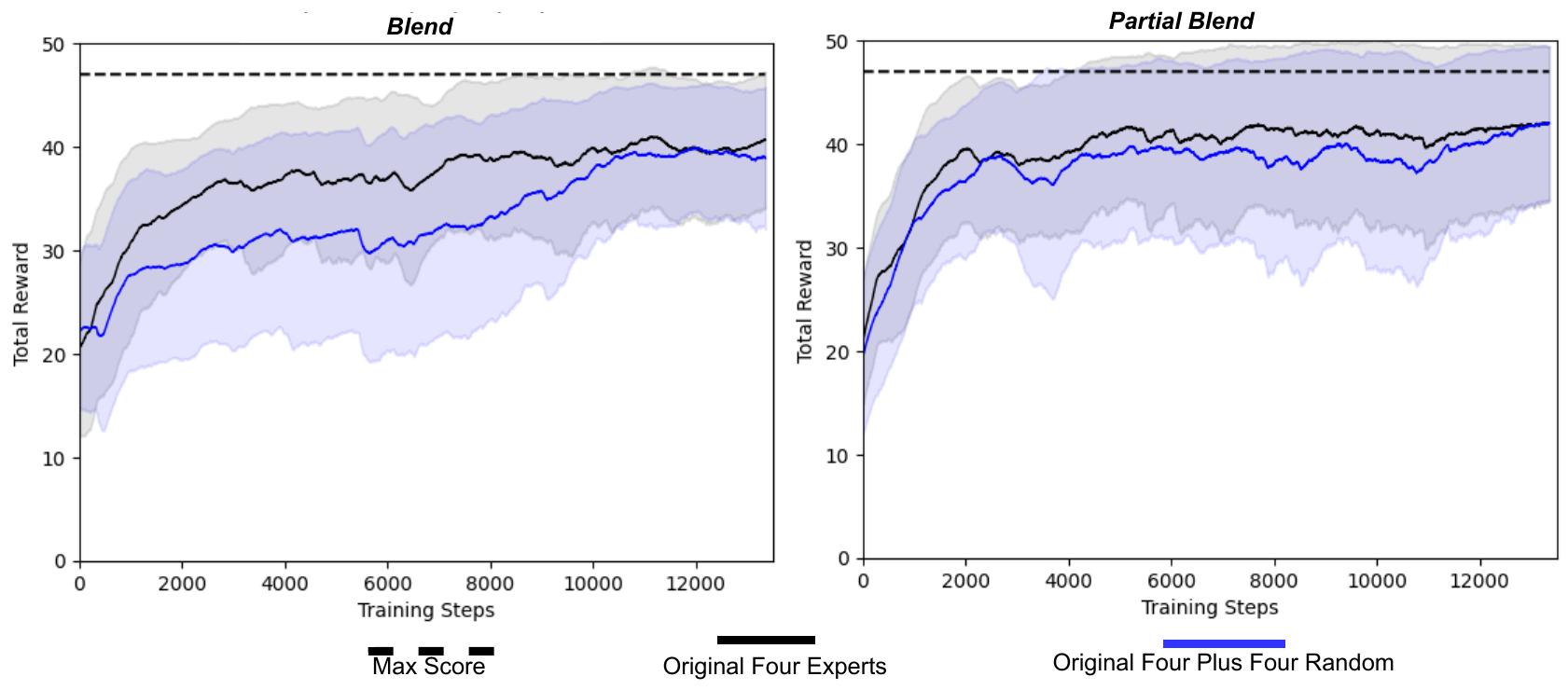}
    \label{fig:random}
    \caption{The testing-time performance of the MoE agent with only the original four experts versus the same agent with four added random experts. When all experts contribute some relevant information to the new task, the MoE agent's performance suffers slightly as the attention module needs more time to distinguish relevant experts from irrelevant experts. When there are only a few relevant experts, additional irrelevant experts have little to no impact on the MoE agent's performance.}
\end{figure}

For the blended target, we see only a marginal drop in early performance due to the noise from the random agents causing the MoE agent to take longer to learn which experts to attend to and thus discover rewards. 
However, performance recovers as the attention module learns to ignore the random agents after an extended period of training. 

Conversely, in the partial blend we see the 4 random agents having virtually no impact on the performance of the MoE agent. While the random agents have a negative impact on the attention module in its ability to identify the most relevant experts for particular state, this becomes less of an issue with a lower number of relevant experts.

\paragraph{Irrelevant Experts.} To examine the MoE agent's robustness to fewer, or irrelevant, experts, we remove the relevant experts to the partial blend and run the MoE agent with only irrelevant experts. 
This can be considered an adverarial attack on the agent, requiring it to perform a task that is completely unlike any that it has ever mastered.
Doing this drastically degrades the MoE agent's performance due to the adversarial nature of this arrangement. 
In a case where no expert has any useful knowledge with respect to the target role, the MoE agent is effectively reduced to training KG-A2C from scratch, but experts want to pull the agent away from promising exploration. 
Even with epsilon-greedy exploration style, the experts will not let the agent explore too far.


\section{Conclusions}
\label{sec: conclusion}
In this paper, we introduce a novel method for task transfer in open-ended, text-based environments. We demonstrate how performing a role can be framed as a task such that a variety of frozen experts, each trained in a specific role, can be mixed by an attention module along with a new, trainable expert to create a MoE agent capable of rapidly learning a new, related task. In this MoE agent, the attention module assigns scores to each expert based on the current observation. Pre-trained experts are more closely attended to in contextually similar states with the new, trainable experts filling in the parts of the policy that cannot be transferred from the pre-trained experts. We show our MoE agent, when not facing adversarial settings, far outperforms any single individual of its parts in a few-shot setting and demonstrate its robustness to a large number of irrelevant experts.


\bibliography{main}

\begin{thebibliography}{44}
\providecommand{\natexlab}[1]{#1}
\providecommand{\url}[1]{\texttt{#1}}
\expandafter\ifx\csname urlstyle\endcsname\relax
  \providecommand{\doi}[1]{doi: #1}\else
  \providecommand{\doi}{doi: \begingroup \urlstyle{rm}\Url}\fi

\bibitem[Abdulhai et~al.(2023)Abdulhai, White, Snell, Sun, Hong, Zhai, Xu, and Levine]{abdulhai2023lmrl}
Marwa Abdulhai, Isadora White, Charlie Snell, Charles Sun, Joey Hong, Yuexiang Zhai, Kelvin Xu, and Sergey Levine.
\newblock Lmrl gym: Benchmarks for multi-turn reinforcement learning with language models, 2023.

\bibitem[Adhikari et~al.(2020)Adhikari, Yuan, C{\^o}t{\'e}, Zelinka, Rondeau, Laroche, Poupart, Tang, Trischler, and Hamilton]{adhikari2020learning}
Ashutosh Adhikari, Xingdi Yuan, Marc-Alexandre C{\^o}t{\'e}, Mikul{\'a}{\v{s}} Zelinka, Marc-Antoine Rondeau, Romain Laroche, Pascal Poupart, Jian Tang, Adam Trischler, and Will Hamilton.
\newblock Learning dynamic belief graphs to generalize on text-based games.
\newblock \emph{Advances in Neural Information Processing Systems}, 33, 2020.

\bibitem[Ammanabrolu et~al.(2021)Ammanabrolu, Urbanek, Li, Szlam, Rockt{\"a}schel, and Weston]{ammanabrolu2021motivate}
P~Ammanabrolu, J~Urbanek, M~Li, A~Szlam, T~Rockt{\"a}schel, and J~Weston.
\newblock How to motivate your dragon: Teaching goal-driven agents to speak and act in fantasy worlds.
\newblock In \emph{NAACL-HLT}. Association for Computational Linguistics, 2021.

\bibitem[Ammanabrolu \& Hausknecht(2019)Ammanabrolu and Hausknecht]{ammanabrolu2019graph}
Prithviraj Ammanabrolu and Matthew Hausknecht.
\newblock Graph constrained reinforcement learning for natural language action spaces.
\newblock In \emph{International Conference on Learning Representations}, 2019.

\bibitem[Ammanabrolu \& Riedl(2019)Ammanabrolu and Riedl]{ammanabrolu2019playing}
Prithviraj Ammanabrolu and Mark Riedl.
\newblock Playing text-adventure games with graph-based deep reinforcement learning.
\newblock In \emph{Proceedings of the 2019 Conference of the North American Chapter of the Association for Computational Linguistics: Human Language Technologies, Volume 1 (Long and Short Papers)}, 2019.

\bibitem[Ammanabrolu \& Riedl(2018)Ammanabrolu and Riedl]{ammanabrolu2018playing}
Prithviraj Ammanabrolu and Mark~O Riedl.
\newblock Playing text-adventure games with graph-based deep reinforcement learning.
\newblock \emph{arXiv preprint arXiv:1812.01628}, 2018.

\bibitem[Ammanabrolu et~al.(2020)Ammanabrolu, Tien, Hausknecht, and Riedl]{ammanabrolu2020avoid}
Prithviraj Ammanabrolu, Ethan Tien, Matthew Hausknecht, and Mark~O Riedl.
\newblock How to avoid being eaten by a grue: Structured exploration strategies for textual worlds.
\newblock \emph{arXiv preprint arXiv:2006.07409}, 2020.

\bibitem[Ba et~al.(2016)Ba, Kiros, and Hinton]{ba2016layer}
Jimmy~Lei Ba, Jamie~Ryan Kiros, and Geoffrey~E Hinton.
\newblock Layer normalization.
\newblock \emph{arXiv preprint arXiv:1607.06450}, 2016.

\bibitem[Bahdanau et~al.(2016)Bahdanau, Brakel, Xu, Goyal, Lowe, Pineau, Courville, and Bengio]{bahdanau2016actor}
Dzmitry Bahdanau, Philemon Brakel, Kelvin Xu, Anirudh Goyal, Ryan Lowe, Joelle Pineau, Aaron Courville, and Yoshua Bengio.
\newblock An actor-critic algorithm for sequence prediction.
\newblock \emph{arXiv preprint arXiv:1607.07086}, 2016.

\bibitem[Brown et~al.(2020)Brown, Mann, Ryder, Subbiah, Kaplan, Dhariwal, Neelakantan, Shyam, Sastry, Askell, et~al.]{brown2020language}
Tom Brown, Benjamin Mann, Nick Ryder, Melanie Subbiah, Jared~D Kaplan, Prafulla Dhariwal, Arvind Neelakantan, Pranav Shyam, Girish Sastry, Amanda Askell, et~al.
\newblock Language models are few-shot learners.
\newblock \emph{Advances in neural information processing systems}, 33, 2020.

\bibitem[Carta et~al.(2023)Carta, Romac, Wolf, Lamprier, Sigaud, and Oudeyer]{carta2023grounding}
Thomas Carta, Clément Romac, Thomas Wolf, Sylvain Lamprier, Olivier Sigaud, and Pierre-Yves Oudeyer.
\newblock Grounding large language models in interactive environments with online reinforcement learning, 2023.

\bibitem[Cheng et~al.(2020)Cheng, Kolobov, and Agarwal]{Cheng2020}
Ching-An Cheng, Andrey Kolobov, and Alekh Agarwal.
\newblock Policy improvement via imitation of multiple oracles.
\newblock In H.~Larochelle, M.~Ranzato, R.~Hadsell, M.F. Balcan, and H.~Lin (eds.), \emph{Advances in Neural Information Processing Systems}, volume~33, pp.\  5587--5598. Curran Associates, Inc., 2020.
\newblock URL \url{https://proceedings.neurips.cc/paper_files/paper/2020/file/3c56fe2f24038c4d22b9eb0aca78f590-Paper.pdf}.

\bibitem[Cheng et~al.(2023)Cheng, Dong, Cai, and Sun]{Cheng2023}
Guangran Cheng, Lu~Dong, Wenzhe Cai, and Changyin Sun.
\newblock Multi-task reinforcement learning with attention-based mixture of experts.
\newblock \emph{IEEE Robotics and Automation Letters}, 8\penalty0 (6):\penalty0 3812--3819, 2023.
\newblock \doi{10.1109/LRA.2023.3271445}.

\bibitem[Chhikara et~al.(2023)Chhikara, Zhang, Ilievski, Francis, and Ma]{Prateek2023}
Prateek Chhikara, Jiarui Zhang, Filip Ilievski, Jonathan Francis, and Kaixin Ma.
\newblock Knowledge-enhanced agents for interactive text games.
\newblock In \emph{Proceedings of the 12th Knowledge Capture Conference 2023}, pp.\  157–165. Association for Computing Machinery, 2023.

\bibitem[Devlin et~al.(2019{\natexlab{a}})Devlin, Chang, Lee, and Toutanova]{devlin-etal-2019-bert}
Jacob Devlin, Ming-Wei Chang, Kenton Lee, and Kristina Toutanova.
\newblock {BERT}: Pre-training of deep bidirectional transformers for language understanding.
\newblock In \emph{Proceedings of the 2019 Conference of the North {A}merican Chapter of the Association for Computational Linguistics: Human Language Technologies, Volume 1 (Long and Short Papers)}. Association for Computational Linguistics, June 2019{\natexlab{a}}.
\newblock \doi{10.18653/v1/N19-1423}.

\bibitem[Devlin et~al.(2019{\natexlab{b}})Devlin, Chang, Lee, and Toutanova]{devlin2019bert}
Jacob Devlin, Ming-Wei Chang, Kenton Lee, and Kristina Toutanova.
\newblock Bert: Pre-training of deep bidirectional transformers for language understanding.
\newblock In \emph{Proceedings of the 2019 Conference of the North American Chapter of the Association for Computational Linguistics: Human Language Technologies, Volume 1 (Long and Short Papers)}, 2019{\natexlab{b}}.

\bibitem[Elfwing et~al.(2018)Elfwing, Uchibe, and Doya]{elfwing2018sigmoid}
S~Elfwing, E~Uchibe, and K~Doya.
\newblock Sigmoid-weighted linear units for neural network function approximation in reinforcement learning.
\newblock \emph{Neural Networks: the Official Journal of the International Neural Network Society}, 107:\penalty0 3--11, 2018.

\bibitem[Fang et~al.(2021)Fang, Zhu, Savarese, and Fei{-}Fei]{Fang2021}
Kuan Fang, Yuke Zhu, Silvio Savarese, and Li~Fei{-}Fei.
\newblock Discovering generalizable skills via automated generation of diverse tasks.
\newblock \emph{CoRR}, abs/2106.13935, 2021.
\newblock URL \url{https://arxiv.org/abs/2106.13935}.

\bibitem[Hausknecht et~al.(2020)Hausknecht, Ammanabrolu, C{\^o}t{\'e}, and Yuan]{hausknecht2020interactive}
Matthew Hausknecht, Prithviraj Ammanabrolu, Marc-Alexandre C{\^o}t{\'e}, and Xingdi Yuan.
\newblock Interactive fiction games: A colossal adventure.
\newblock In \emph{Proceedings of the AAAI Conference on Artificial Intelligence}, volume~34, 2020.

\bibitem[Hinton et~al.(2015)Hinton, Vinyals, and Dean]{hinton2015distilling}
Geoffrey Hinton, Oriol Vinyals, and Jeff Dean.
\newblock Distilling the knowledge in a neural network, 2015.

\bibitem[Hu et~al.(2021)Hu, Wallis, Allen-Zhu, Li, Wang, Wang, Chen, et~al.]{hu2021lora}
Edward~J Hu, Phillip Wallis, Zeyuan Allen-Zhu, Yuanzhi Li, Shean Wang, Lu~Wang, Weizhu Chen, et~al.
\newblock Lora: Low-rank adaptation of large language models.
\newblock In \emph{International Conference on Learning Representations}, 2021.

\bibitem[Kessler~Faulkner \& Thomaz(2021)Kessler~Faulkner and Thomaz]{Faulkner2021}
Taylor~A. Kessler~Faulkner and Andrea Thomaz.
\newblock Interactive reinforcement learning from imperfect teachers.
\newblock HRI '21 Companion, pp.\  577–579, New York, NY, USA, 2021. Association for Computing Machinery.
\newblock ISBN 9781450382908.
\newblock \doi{10.1145/3434074.3446361}.
\newblock URL \url{https://doi.org/10.1145/3434074.3446361}.

\bibitem[Kurenkov et~al.(2020)Kurenkov, Mandlekar, Martin-Martin, Savarese, and Garg]{kurenkov2020ac}
Andrey Kurenkov, Ajay Mandlekar, Roberto Martin-Martin, Silvio Savarese, and Animesh Garg.
\newblock Ac-teach: A bayesian actor-critic method for policy learning with an ensemble of suboptimal teachers.
\newblock In \emph{Conference on Robot Learning}, pp.\  717--734. PMLR, 2020.

\bibitem[Li et~al.(2023)Li, Li, Zhang, Wang, Liu, and Zhang]{li2023iob}
Siyuan Li, Hao Li, Jin Zhang, Zhen Wang, Peng Liu, and Chongjie Zhang.
\newblock Iob: Integrating optimization transfer and behavior transfer for multi-policy reuse, 2023.

\bibitem[Mnih et~al.(2016)Mnih, Badia, Mirza, Graves, Lillicrap, Harley, Silver, and Kavukcuoglu]{mnih2016asynchronous}
Volodymyr Mnih, Adria~Puigdomenech Badia, Mehdi Mirza, Alex Graves, Timothy Lillicrap, Tim Harley, David Silver, and Koray Kavukcuoglu.
\newblock Asynchronous methods for deep reinforcement learning.
\newblock In \emph{International conference on machine learning}. PMLR, 2016.

\bibitem[Murugesan et~al.(2020)Murugesan, Atzeni, Shukla, Sachan, Kapanipathi, and Talamadupula]{murugesan2020enhancing}
Keerthiram Murugesan, Mattia Atzeni, Pushkar Shukla, Mrinmaya Sachan, Pavan Kapanipathi, and Kartik Talamadupula.
\newblock Enhancing text-based reinforcement learning agents with commonsense knowledge.
\newblock \emph{arXiv preprint arXiv:2005.00811}, 2020.

\bibitem[Narasimhan et~al.(2015)Narasimhan, Kulkarni, and Barzilay]{narasimhan2015language}
Karthik Narasimhan, Tejas Kulkarni, and Regina Barzilay.
\newblock Language understanding for text-based games using deep reinforcement learning.
\newblock In \emph{Proceedings of the 2015 Conference on Empirical Methods in Natural Language Processing}, 2015.

\bibitem[Pan et~al.(2023)Pan, Chan, Zou, Li, Basart, Woodside, Ng, Zhang, Emmons, and Hendrycks]{pan2023rewards}
Alexander Pan, Jun~Shern Chan, Andy Zou, Nathaniel Li, Steven Basart, Thomas Woodside, Jonathan Ng, Hanlin Zhang, Scott Emmons, and Dan Hendrycks.
\newblock Do the rewards justify the means? measuring trade-offs between rewards and ethical behavior in the machiavelli benchmark, 2023.

\bibitem[Parisotto et~al.(2016)Parisotto, Ba, and Salakhutdinov]{parisotto2016actormimic}
Emilio Parisotto, Jimmy~Lei Ba, and Ruslan Salakhutdinov.
\newblock Actor-mimic: Deep multitask and transfer reinforcement learning, 2016.

\bibitem[Peng et~al.(2023)Peng, Cui, Zhou, Jia, and Riedl]{peng2023story}
Xiangyu Peng, Christopher Cui, Wei Zhou, Renee Jia, and Mark Riedl.
\newblock Story shaping: Teaching agents human-like behavior with stories.
\newblock \emph{arXiv preprint arXiv:2301.10107}, 2023.

\bibitem[Prasad et~al.(2023)Prasad, Koller, Hartmann, Clark, Sabharwal, Bansal, and Khot]{prasad2023adapt}
Archiki Prasad, Alexander Koller, Mareike Hartmann, Peter Clark, Ashish Sabharwal, Mohit Bansal, and Tushar Khot.
\newblock Adapt: As-needed decomposition and planning with language models, 2023.

\bibitem[Raffel et~al.(2020)Raffel, Shazeer, Roberts, Lee, Narang, Matena, Zhou, Li, and Liu]{raffel2020exploring}
Colin Raffel, Noam Shazeer, Adam Roberts, Katherine Lee, Sharan Narang, Michael Matena, Yanqi Zhou, Wei Li, and Peter~J Liu.
\newblock Exploring the limits of transfer learning with a unified text-to-text transformer.
\newblock \emph{The Journal of Machine Learning Research}, 21\penalty0 (1), 2020.

\bibitem[Rusu et~al.(2016)Rusu, Colmenarejo, Gulcehre, Desjardins, Kirkpatrick, Pascanu, Mnih, Kavukcuoglu, and Hadsell]{rusu2016policy}
Andrei~A. Rusu, Sergio~Gomez Colmenarejo, Caglar Gulcehre, Guillaume Desjardins, James Kirkpatrick, Razvan Pascanu, Volodymyr Mnih, Koray Kavukcuoglu, and Raia Hadsell.
\newblock Policy distillation, 2016.

\bibitem[Ryu et~al.(2023)Ryu, Fang, Haffari, Pan, and Shareghi]{ryu-etal-2023-minimal}
Dongwon Ryu, Meng Fang, Gholamreza Haffari, Shirui Pan, and Ehsan Shareghi.
\newblock A minimal approach for natural language action space in text-based games.
\newblock In Jing Jiang, David Reitter, and Shumin Deng (eds.), \emph{Proceedings of the 27th Conference on Computational Natural Language Learning (CoNLL)}, pp.\  138--154, Singapore, December 2023. Association for Computational Linguistics.

\bibitem[Schmitt et~al.(2018)Schmitt, Hudson, Zidek, Osindero, Doersch, Czarnecki, Leibo, Kuttler, Zisserman, Simonyan, and Eslami]{schmitt2018kickstarting}
Simon Schmitt, Jonathan~J. Hudson, Augustin Zidek, Simon Osindero, Carl Doersch, Wojciech~M. Czarnecki, Joel~Z. Leibo, Heinrich Kuttler, Andrew Zisserman, Karen Simonyan, and S.~M.~Ali Eslami.
\newblock Kickstarting deep reinforcement learning, 2018.

\bibitem[Schulman et~al.(2018)Schulman, Chen, and Abbeel]{schulman2018equivalence}
John Schulman, Xi~Chen, and Pieter Abbeel.
\newblock Equivalence between policy gradients and soft q-learning, 2018.

\bibitem[Shridhar et~al.(2021)Shridhar, Yuan, C\^ot\'e, Bisk, Trischler, and Hausknecht]{ALFWorld20}
Mohit Shridhar, Xingdi Yuan, Marc-Alexandre C\^ot\'e, Yonatan Bisk, Adam Trischler, and Matthew Hausknecht.
\newblock {ALFWorld: Aligning Text and Embodied Environments for Interactive Learning}.
\newblock In \emph{Proceedings of the International Conference on Learning Representations (ICLR)}, 2021.
\newblock URL \url{https://arxiv.org/abs/2010.03768}.

\bibitem[Sodhani et~al.(2021)Sodhani, Zhang, and Pineau]{sodhani2021multitask}
Shagun Sodhani, Amy Zhang, and Joelle Pineau.
\newblock Multi-task reinforcement learning with context-based representations, 2021.

\bibitem[Teh et~al.(2017)Teh, Bapst, Czarnecki, Quan, Kirkpatrick, Hadsell, Heess, and Pascanu]{Distral2017}
Yee~Whye Teh, Victor Bapst, Wojciech~Marian Czarnecki, John Quan, James Kirkpatrick, Raia Hadsell, Nicolas Heess, and Razvan Pascanu.
\newblock Distral: Robust multitask reinforcement learning.
\newblock \emph{CoRR}, abs/1707.04175, 2017.
\newblock URL \url{http://arxiv.org/abs/1707.04175}.

\bibitem[Urbanek et~al.(2019)Urbanek, Fan, Karamcheti, Jain, Humeau, Dinan, Rockt{\"a}schel, Kiela, Szlam, and Weston]{urbanek2019learning}
Jack Urbanek, Angela Fan, Siddharth Karamcheti, Saachi Jain, Samuel Humeau, Emily Dinan, Tim Rockt{\"a}schel, Douwe Kiela, Arthur Szlam, and Jason Weston.
\newblock Learning to speak and act in a fantasy text adventure game.
\newblock In \emph{Proceedings of the 2019 Conference on Empirical Methods in Natural Language Processing and the 9th International Joint Conference on Natural Language Processing (EMNLP-IJCNLP)}, 2019.

\bibitem[Veličković et~al.(2018)Veličković, Cucurull, Casanova, Romero, Liò, and Bengio]{veličković2018graph}
Petar Veličković, Guillem Cucurull, Arantxa Casanova, Adriana Romero, Pietro Liò, and Yoshua Bengio.
\newblock Graph attention networks.
\newblock In \emph{International Conference on Learning Representations}, 2018.
\newblock URL \url{https://openreview.net/forum?id=rJXMpikCZ}.

\bibitem[Wang et~al.(2022)Wang, Jansen, Côté, and Ammanabrolu]{Wang2022Science}
Ruoyao Wang, Peter Jansen, Marc-Alexandre Côté, and Prithviraj Ammanabrolu.
\newblock Scienceworld: Is your agent smarter than a 5th grader?, 2022.
\newblock URL \url{https://arxiv.org/abs/2203.07540}.

\bibitem[Wang et~al.(2023)Wang, Jansen, C{\^o}t{\'e}, and Ammanabrolu]{wang2023behavior}
Ruoyao Wang, Peter Jansen, Marc-Alexandre C{\^o}t{\'e}, and Prithviraj Ammanabrolu.
\newblock Behavior cloned transformers are neurosymbolic reasoners.
\newblock In \emph{Proceedings of the 17th Conference of the European Chapter of the Association for Computational Linguistics}, pp.\  2769--2780, 2023.

\bibitem[Yin \& Pan(2017)Yin and Pan]{Yin2017}
Haiyan Yin and Sinno~Jialin Pan.
\newblock Knowledge transfer for deep reinforcement learning with hierarchical experience replay.
\newblock AAAI'17, pp.\  1640–1646. AAAI Press, 2017.

\end{thebibliography}
\bibliographystyle{rlc}

\newpage
\appendix
\section*{Appendix}
\appendix
\section{KG-A2C}\label{KG-A2C}
For our source agents, we build off the KG-A2C agent framework~\cite{ammanabrolu2019graph}, an Advantage-Actor Critic architecture augmented with a knowledge-graph based attention. 
KG-A2C's space of observations 
includes (a)~text description of the room the agent is in via the ``look'' command,
(b)~text descriptions of the character's inventory via the ``inventory'' command,
(c)~feedback from the last command, and
(d)~the agent's last command.
The state observations are concatenated and embedded using a recurrent GRU.

Simultaneously, the state observation is used to update a knowledge graph of facts about the world that have been observed to date.
This includes facts and relations about rooms, objects in rooms, inventory items, etc.
This knowledge graph is then embedded using a graph attention mechanism~\cite{veličković2018graph}. 

Advantage-actor critic networks \cite{mnih2016asynchronous} have two heads. 
The actor head generates logit scores, one for each possible action, which can be converted to a probability distribution via softmax and sampled to determine which action the agent takes.
The critic head estimates the utility of the state.
Actions are made up of verbs and optional object names.
The KG-A2C agent generates a verb, which maps to a pre-defined template, and the generated object name is used to populate the template. 

\subsection{Game Map}\label{app:game_map}
\subsubsection{Game Map Part A}
\includegraphics[width=\textwidth]{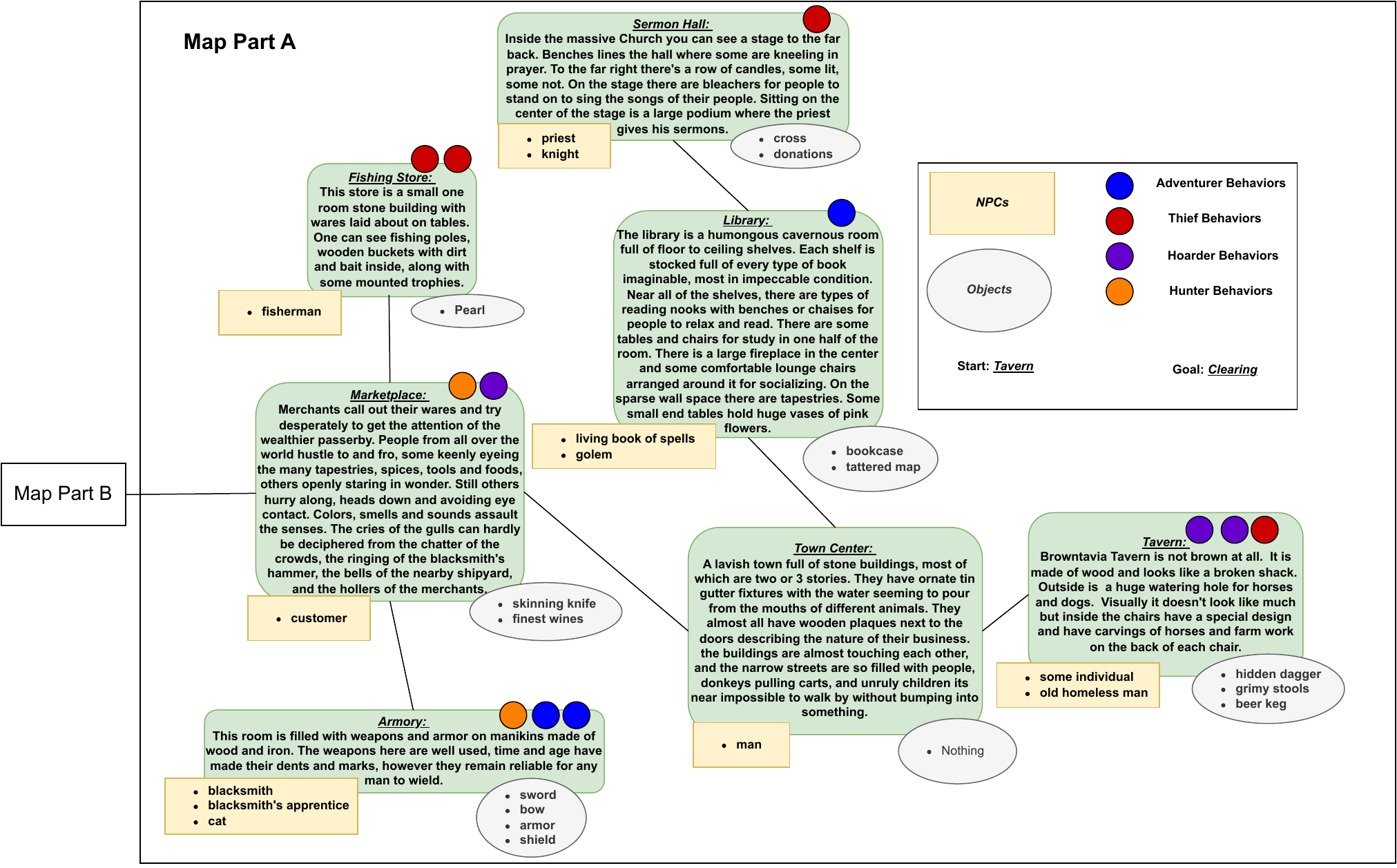}

\subsubsection{Game Map Part B}
\includegraphics[width=\textwidth]{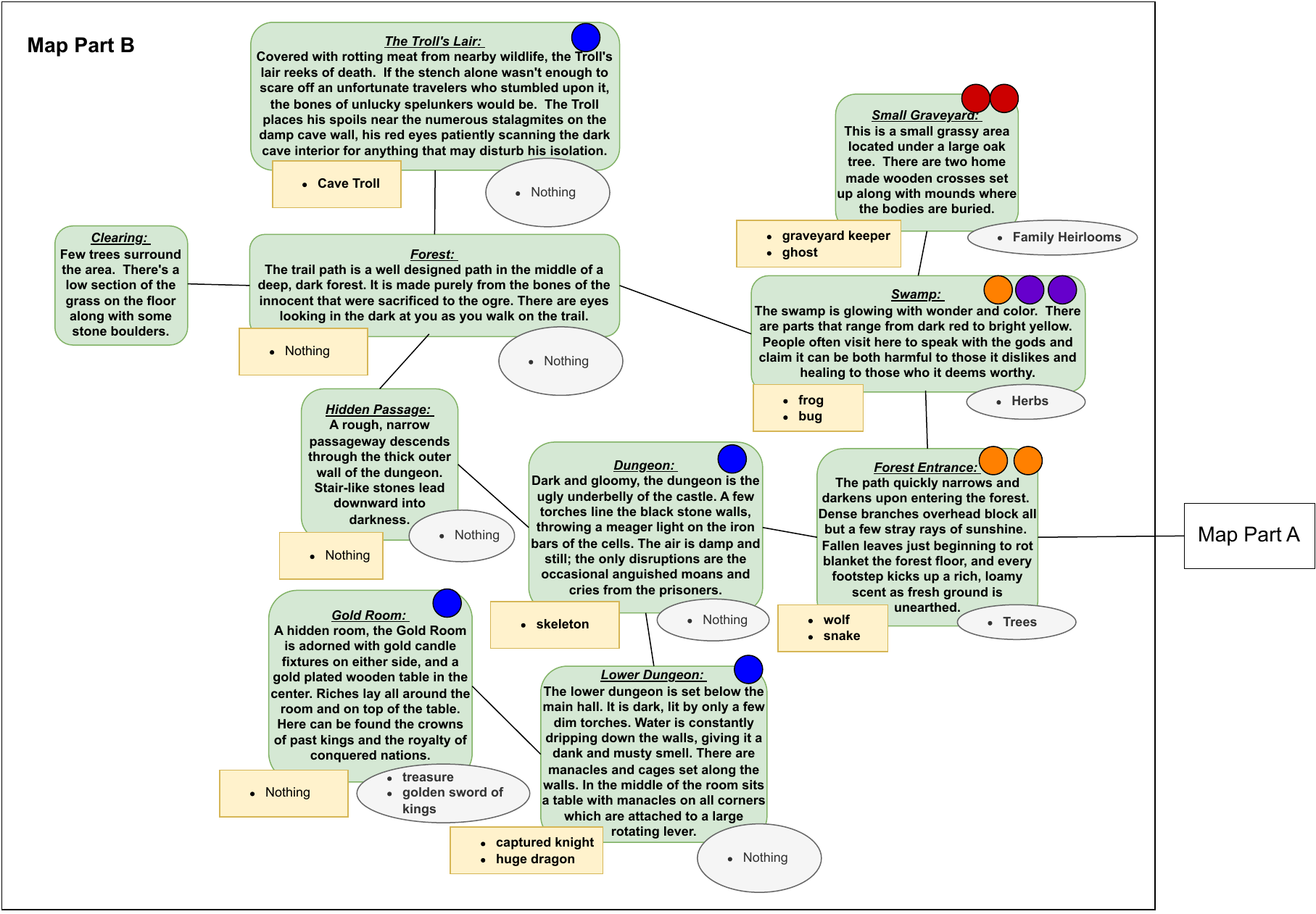}

\section{Story Shaping}\label{Story Shaping}
We train all KG-A2C models with Story Shaping ~\cite{peng2023story}. The following are the stories used to align the pre-trained agents with their respective roles.

\paragraph{Adventurer.} I am an adventurer. I have come to this town seeking treasure and challenge. In the library, I find a tattered map that tells of a dungeon and the treasure within. I first go to the armory to purchase a sword and armor. Leaving town, I make my way to the dungeon. I slay the skeleton at the entrance and travel deeper into the dungeon. In the lower dungeon, I encounter a huge dragon. After an arduous battle, I slay the huge dragon. Past the dragon, I find a gold room and take the treasure. On the way to the clearing, I note the bodies on my path and find a Troll's Lair. Within, I slay the cave troll. My thirst for battle and treasure sated, I travel to the Clearing and set off for the next town. 

\paragraph{Thief.} I am a thief. I value stealth and wealth, only getting into a fight when I have the advantage. In the tavern, I find a hidden dagger and take it for my own. I explore the town, looking for easy targets. In the sermon hall, I steal the donations but avoid the priest due to the knight. In the marketplace, I find a fishing store with a lone fisherman. I kill the fisherman and steal the pearl. Fleeing town, I avoid the dungeon and take a detour through the swamp. In the swamp, I see a small graveyard with single graveyard keeper and some graves. I stab the graveyard keeper in the back, killing him instantly. I dig up the graves and take the valuable family heirlooms. I travel to the Clearing and make my escape.

\paragraph{Hunter.} I am a hunter. My trade is in navigating natural environments and dealing with wildlife. I buy a skinning knife from the markets and get a bow from the armory before I set off. As I travel through the forest, I deal with the wildlife as they come. I kill and collect a wolf and a snake. I opt to take the route through the swamps rather than the dungeon, due to my comfort with nature. I encounter and shoot a poisonous frog but otherwise face no issue as I travel to the Clearing and make my way to the next town.

\paragraph{Hoarder.} I am a hoarder. I love to collect useless things and anything I can get my hands on. I get a drink from the tavern but keep the beer keg. The owner chases me out but I grab the grimy bar stools as I go. In the marketplace, I can't control myself and take the finest wines. I am promptly chased out of town by the guards. Feeling hungry, I grab some herbs to tide me over. With no weapons, I make sure to avoid the dungeon and instead travel through the swamp, accidentally stepping on a bug. I exit through the clearing with my strange collection of items.

The corresponding triples generated for target knowledge graphs $G_1, G_2, G_3$ and $G_4$ result in the following role-aligned behaviors:

\paragraph{Adventurer Nodes.} $get\_sword, get\_armor, get\_tattered\_map, hit\_skeleton, hit\_huge\_dragon,$
$ get\_treasure, hit\_cave\_troll$

\paragraph{Thief Nodes.} $get\_hidden\_dagger, get\_donations, get\_pearl, hit\_fisherman, hit\_graveyard\_keeper,$ $ get\_family\_heirlooms$

\paragraph{Hunter Nodes.} $get\_bow^*, get\_skinning\_knife, hit\_snake, hit\_wolf, hit\_frog$

\paragraph{Hoarder Nodes.} $get\_beer\_keg, get\_grimy\_stools, get\_finest\_wines, get\_herbs, hit\_bug$

* indicates the respective expert did not successfully obtain this reward during training. 

\section{Best Models}
The following table contains the average total score and steps of the best MoE agent for each of the 5 seeds for all target roles. Models were ranked in order of highest average total score, number of steps and total training steps.

\begin{table}[h]
    \centering
    \begin{tabular}{|c|c|c|c|c||c|c|c|c|c||c|}
    \hline
    \multicolumn{11}{|c|}{Blend} \\
        \hline
        \multicolumn{1}{|c||}{Metric} &
        \multicolumn{5}{|c||}{Target Role 1} & \multicolumn{5}{c||}{Target Role 2} \\
        \hline
        Avg Score & 47.0 & 47.0 & \textbf{34.5} & 47.0 & 47.0 & 47.0 & 47.0 & 47.0 & 47.0 & 47.0\\
        Avg Steps & 22.875 & 25.375 & \textbf{18.666} & 22.5 & 22.375 & 18.0 & 18.0 & 22.0 & 18.0 & 18.0\\
        Training Steps & 8912 & 9331 & \textbf{12963} & 11773 & 6725 & 11817 & 8144 & 13245 & 5761 & 10832\\
        \hline
        \multicolumn{11}{|c|}{Partial Blend} \\
        \hline
        \multicolumn{1}{|c||}{Metric} &
        \multicolumn{5}{|c||}{Target Role 3} & \multicolumn{5}{c||}{Target Role 4}\\
        \hline
        Avg Score & 47.0 & 47.0 & 47.0 & 47.0 & 47.0 & 47.0 & 47.0 & 47.0 & 47.0 & 47.0\\
        Avg Steps & 26.125 & 24.125 & 25.5 & 27.625 & 27.375 & 19.143 &19.75 & 21 & 19.0 & 18.75 \\
        Training Steps & 11804 & 10887 & 8199 & 11330 & 10073 & 11615 & 10055 & 10458 & 8269 & 9356 \\
        \hline
    \end{tabular}
    \caption{Average score and steps of all best MoE agents across ten games after training. Target Role 1 and 4 are those shown in \ref{sec:results}. Our epsilon-greedy coefficient only decays to 0 after 10000 training steps. Despite this, only one seed fails to produce a model that can achieve the max score across 10 testing time games.}
    \label{tab:my_label}
\end{table}

\end{document}